\documentclass[letterpaper, 10 pt, conference]{ieeeconf}
\usepackage{graphics} 
\usepackage{epsfig}   
\usepackage{mathptmx} 
\usepackage{times}    
\usepackage{amsmath}  
\usepackage{amssymb}  
\usepackage{color}
\usepackage{verbatim}
\title{\LARGE \bf
Controllable Mechanical-domain Energy Accumulators}

\author{Sung Y. Kim and David J. Braun
	\thanks{
	Sung Y. Kim and David J. Braun are with the Advanced Robotics and Control Laboratory within the Center for Rehabilitation Engineering and Assistive Technology, Department of Mechanical Engineering, Vanderbilt University, Nashville, TN 37235. \newline
	\indent This work was supported in part by a Seeding Success Grant provide by Vanderbilt University and a National Science Foundation CAREER Award (Grant No. 2144551). The authors gratefully acknowledge the support.
	\newline
	\indent This paper has supplementary downloadable multimedia material available at http://ieeexplore.ieee.org. The video demonstrates the locking of the capstan clutch mechanism.}
	\thanks{Email: {\tt\small sung.kim@vanderbilt.edu}}
	\thanks{Email: {\tt\small david.braun@vanderbilt.edu}\newline
		\indent © 2023 IEEE. Personal use of this material is permitted. Permission from IEEE must be obtained for all other uses, in any current or future media, including reprinting/republishing this material for advertising or promotional purposes, creating new collective works, for resale or redistribution to servers or lists, or reuse of any copyrighted component of this work in other works.}
}
\IEEEoverridecommandlockouts
\begin{document}
\maketitle
\thispagestyle{empty}
\pagestyle{empty}
\begin{abstract}
Springs are efficient in storing and returning elastic potential energy but are unable to hold the energy they store in the absence of an external load. Lockable springs use clutches to hold elastic potential energy in the absence of an external load, but have not yet been widely adopted in applications, partly because clutches introduce design complexity, reduce energy efficiency, and typically do not afford high fidelity control over the energy stored by the spring. Here, we present the design of a novel lockable compression spring that uses a small capstan clutch to passively lock a mechanical spring. The capstan clutch can lock over $1000~N$ force at any arbitrary deflection, unlock the spring in less than $10~ms$ with a control force less than $1\%$ of the maximal spring force, and provide an $80\%$ energy storage and return efficiency (comparable to a highly efficient electric motor operated at constant nominal speed). By retaining the form factor of a regular spring while providing high-fidelity locking capability even under large spring forces, the proposed design could facilitate the development of energy-efficient spring-based actuators and robots.
\end{abstract}
\section{Introduction}
The use of springs has shown significant energetic benefits in a range of applications where the ability of a spring to store and release energy can assist motor-driven and human-driven robots. 
Springs are especially useful in creating energy efficient oscillatory motion by resonance based actuation \cite{Braun2011, Mathews2021, Balderas2022} as key components of compliant actuators \cite{Braun2019,Braun2019b}, prosthetic devices \cite{Collins2005, Rouse2014,Grimmer2016,Gao2018} and human exoskeletons developed for lifting \cite{Lamers2018,Alemi2019}, walking \cite{Sawicki2009,Collins2015}, and running \cite{Nasiri2018, Simpson2019}. In these applications, springs lower the peak power and torque that must be provided by a motor or human. However, the use of regular springs is limited, as they are unable to hold energy in the absence of an external mechanical load. Adding an internal locking mechanism to a regular spring allows the flow of energy through the spring to be controlled \cite{Plooij2017}, and, as such, 
may alleviate the above-mentioned limitation.
	
Lockable springs can provide a means of controlling the timing of storage and release of energy in spring-driven robots and compliant actuators. When used in parallel with a motor, lockable springs have been shown to reduce peak torque and energy \cite{Haeufle2012, Plooij2016, Heremans2019} by providing precisely timed force assistance. When used in parallel with the ankle joint, springs that lock using latches have been shown to reduce the metabolic cost of walking \cite{Collins2015} by providing force assistance during push-off. Although all of these applications benefit from a spring which can store energy, current lockable spring designs face limitations because clutches introduce design complexity, add mass, reduce energy efficiency, and typically do not afford high fidelity control over the energy stored by the spring. 

In this paper, we present a novel lockable compression spring that uses a small capstan clutch to passively lock a mechanical spring. 
The spring can store over $50~J$ of energy while it enables fast locking and unlocking ($10~ms$) under high loads ($>1000~N$) using small amounts of energy ($0.1~J$), making it advantageous for autonomous applications. 
By providing automatic unidirectional locking when compressed, the spring is able to accumulate energy when compressed multiple times. Furthermore, no external electrical power is required to automatically lock the spring, and minimal power is required to unlock the spring. Because the design closely resembles a regular spring in size and form factor, it can replace regular compression springs without requiring compensatory design adaptations.
		
The proposed lockable spring is a mechanical domain energy accumulator that allows sequential compression and efficient accumulation of elastic potential energy. This kind of spring could be used to design novel compliant actuators \cite{Zhang2021,Kim2021} that could change their force-deflection behavior without changing the energy stored by the spring, useful in physically demanding tasks -- for example, when a human or robot must overcome load inertia during weight-lifting \cite{delooze1993,Toxiri2019,Braun2019b} or when a human or robot must interact with the environment during running \cite{Hunter2005, Sutrisno2020, Zhang2021} and jumping \cite{Bobbert1996, Sutrisno2019}. Furthermore, lockable springs can also find their applications in reducing the energy requirement of industrial pick-and-place robots during gravity compensation, or when the desired motion is periodic but must stop while the end-effector picks up the load. Finally, a lockable spring can provide energetically conservative holding torque without requiring large amounts of electrical power at the motorized joint, and return stored spring potential energy to move the robot with controllable timing at higher actuation efficiency \cite{Plooij2016}. Lockable springs could find their place in next generation actuators and autonomous robots.

\section{Lockable springs} \label{sect: design}
\begin{figure}
	\centering
	\includegraphics[width =\linewidth]{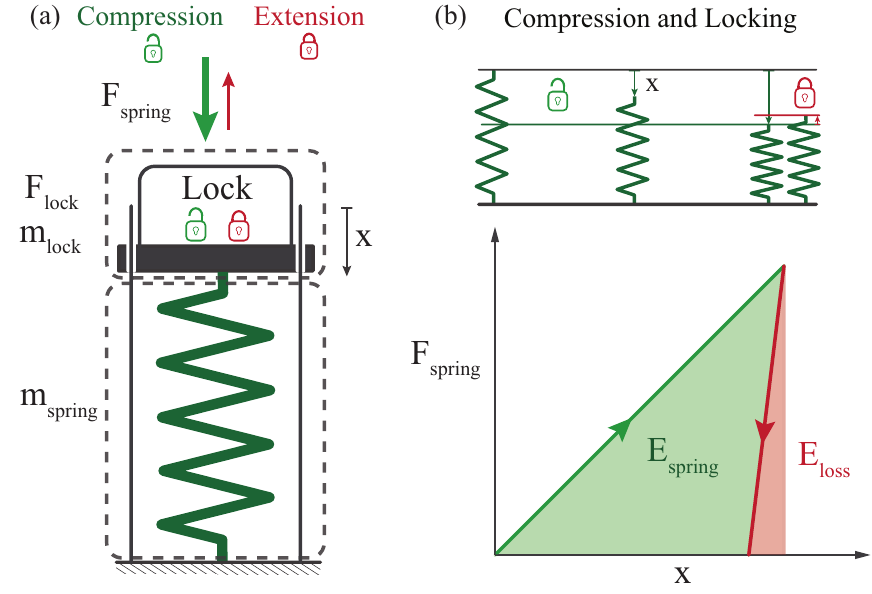}
	\caption{(a) Simple model of a lockable spring. (b) Spring force and elastic potential energy stored during compression and energy loss when locking the length of the spring.}
	\label{fig:simple_model} 
\end{figure}

The ideal lockable spring should retain the characteristics of a regular spring and provide control over the energy storage and release by locking and unlocking the spring. In order to quantify how a design of a lockable spring meets the characteristics of an ideal spring, we will compare (i) the energy efficiency of the lockable spring compared to the energy efficiency of a regular spring, (ii) the force required to lock or unlock the spring compared to the maximal spring force, and the (iii) mass-energy density of the lockable versus the non-lockable spring. These three quantities are introduced below, and a simple representation of a lockable spring is shown in Fig.~\ref{fig:simple_model}(a).

\subsection{Efficiency of locking}
While a regular spring is highly efficient in returning stored potential energy, the added locking function in a lockable spring may introduce energy dissipation and consequently compromise the energy efficiency of the spring. In particular, since the lock cannot engage instantaneously or be completely rigid, there will be energy losses associated with the engagement and disengagement of the lock, see Fig.~\ref{fig:simple_model}(b). In order to quantify the energy loss due to the locking of the spring, we define the energy efficiency of the spring:
\begin{equation}
	\eta = 1- \frac{E_\text{loss}}{E_\text{spring}}\leq 1,
\end{equation}
where $E_\text{loss}$ is the energy loss incurred by the lock in a work-loop where the spring is compressed to store energy, $E_\text{spring}$, prior to the engagement of the lock. 

In applications, $\eta$ should be maximized such that the energy that can be stored and returned by the lockable spring is the same as the energy that can be stored and returned by the spring without the locking mechanism. 
A lockable spring with efficiency $\eta \approx 0.8$ would be similar to a $90\%$ efficient electric motor operated at a constant nominal speed -- first to regenerate energy, and then to drive a robot using the regenerated energy with a $0.9\times 0.9\times 100\% = 81\%$ overall efficiency. 

\begin{figure*}
	\centering
	\includegraphics[width =\textwidth]{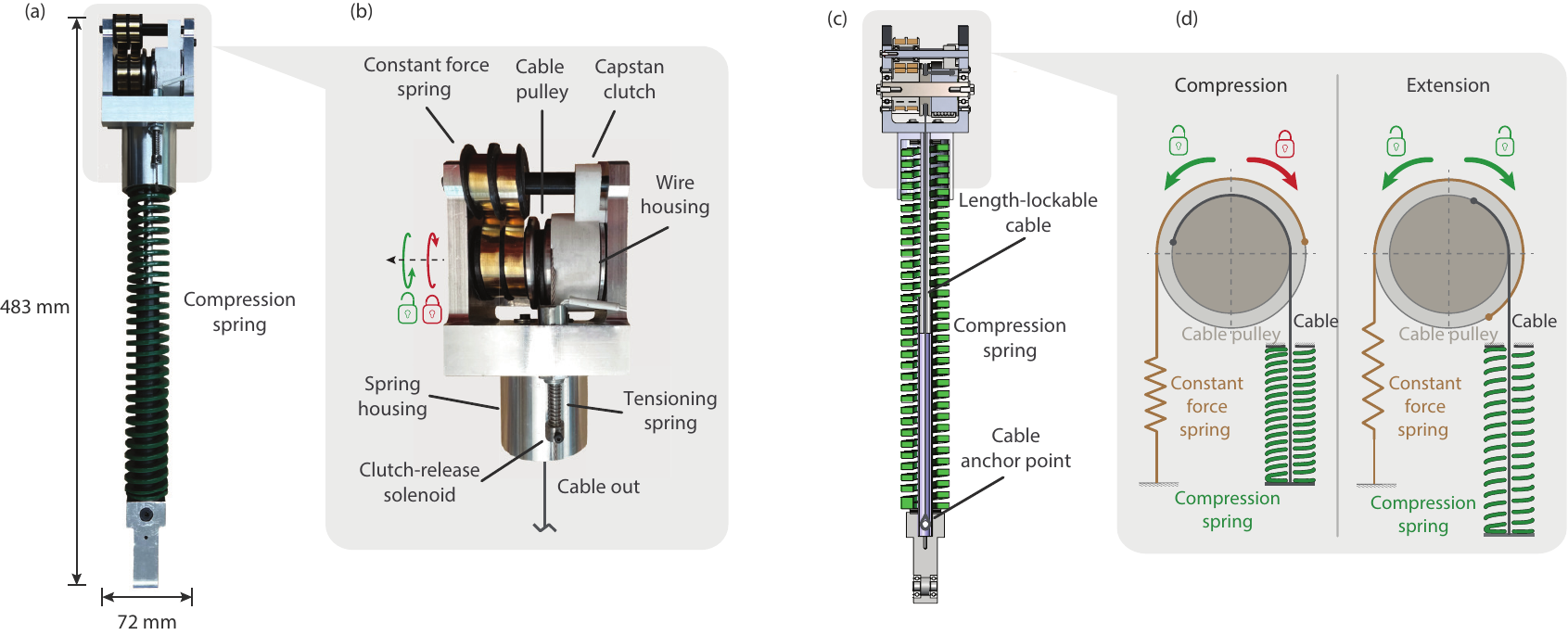}
	\caption{(a)-(b) Lockable compression spring prototype with clutch unit (gray) and compression die spring (green). The spring is always unlocked in compression. The spring can be controlled to lock or unlock in extension. (c) Cross-section of the lockable spring. (d) Working principle of the lockable spring. The small constant force spring (orange) winds the cable pulley when the large spring (green) compresses, while the capstan clutch enables or prevents the rotation of the pulley when the large spring extends.}
	\label{fig:lock_spring}
\end{figure*}

\subsection{Locking and unlocking force}
Ideally, the locking mechanism should easily lock and unlock the spring even if the spring force is large. Locking and unlocking under high loads may be achieved by providing a mechanical advantage to the locking force with a large lever mechanism \cite{Rouse2014}. However, the overall size of the locking mechanism should be small relative to the spring in order to reduce weight. Due to these conflicting objectives, creating small locking mechanisms that can lock and unlock large spring forces has been nontrivial \cite{Plooij2015}.
 
In order to quantify the locking force, we define the ratio between the maximal locking force, $F_\text{lock}$, and the maximal spring force, $F_\text{spring}$, as a function of the non-dimensional design parameters $\Pi$ that define the size -- for example, the maximal width, height, length -- of the locking mechanism: 
\begin{equation}
0\leq \lambda_F = \frac{|F_\text{lock}|}{|F_\text{spring}|}=f(\Pi).
\end{equation}

In applications, $\lambda_F$ should be minimized while taking into account the limited size of the locking mechanism. Our target value is $\lambda_F \leq 0.01$, such that the force required to lock and unlock the spring is at most $1\%$ of the spring force.

\subsection{Mass-energy density}
The mass-energy density of a spring is the amount of elastic potential energy that can be stored in the spring per unit mass. In general, the addition of a locking mechanism increases the total mass of the spring assembly \cite{Plooij2015}, but does not affect the energy storage capacity of the spring. As a result, the added mass of the locking mechanism decreases the mass-energy density of the spring.

The mass-energy density of a lockable spring, $E_\text{spring}/(m_\text{spring}+m_\text{lock})$, compared to a regular, non-lockable spring, $E_\text{spring}/m_\text{spring}$, depends on the mass of the spring and the combined mass of the spring and locking mechanism:
\begin{equation}
	\rho_E = \frac{m_\text{spring}}{m_\text{spring}+m_\text{lock}}\leq 1.
\end{equation} 

In applications, the mass energy density $\rho_E$ should be maximized by minimizing the mass of the locking mechanism. An appropriate target value could be $\rho_E \geq 0.5$, which implies that the added mass of the locking mechanism does not exceed the mass of the spring. 

While an ideal non-lockable spring would be characterized with $\eta=1$, $\lambda_F=0$, and $\rho_E=1$, a practical lockable spring, satisfying $\eta\geq0.8$ ($80\%$ efficiency), $\lambda_F\leq0.01$ ($1\%$ locking force), and $\rho_E\geq 0.5$ ($50\%$ energy density), would enable controllable energy storage and release. 

We note that the proposed metrics can be used to assess differences between designs created for the same purpose. For example, they can be used to compare two different lockable springs that use the same spring but different clutches or compare a lockable spring design to a non-lockable spring design if both designs use the same spring. On the other hand, these metrics should not be used to compare lockable spring designs created for different applications, especially if they use different springs.
\section{Design of a lockable spring}
In this section, we introduce a lockable compression spring where the locking is achieved using a capstan clutch. A prototype comprised of two main components -- a compression spring and a clutch unit -- is shown in Fig.~\ref{fig:lock_spring}(a)-(b). The CAD model of the prototype is shown in Fig.~\ref{fig:lock_spring}(c). The working principle of the spring is illustrated in Fig.~\ref{fig:lock_spring}(d). In what follows, we describe the design and working principle of the lockable compression spring.

\subsection{Prototype}
The prototype uses a compression die spring as the energy storage element, Fig.~\ref{fig:lock_spring}(a,c). The stiffness of the die spring is $14.8$~N/mm, and the length of the spring is $305$~mm. The spring can be compressed $90$~mm to provide over $1200$~N force while storing over $50$~J of elastic potential energy. As the spring is compressed, hollow telescoping shafts, placed within the inner diameter of the spring, prevent buckling of the spring.

The key component of the prototype is the clutch unit, which houses a steel pulley, a constant force spring, and a capstan clutch mechanism. These three components are placed co-axially on a keyed shaft supported in the aluminum housing with bearings, see Fig.~\ref{fig:lock_spring}(b,c). The steel pulley of drum diameter $24$~mm holds a flexible cable that runs through the center of the spring. A braided Technora cable ($1.75$~mm wide, $1.2$~mm thick) was chosen due to its superior flexibility, low elongation under load ($<4\%$), and high strength ($2.7$~kN). The cable is anchored at the bottom of the spring, routed up through the telescoping shafts, wrapped around the pulley, and terminated inside the braking drum of the capstan clutch. Tension in the cable is maintained as the spring is compressed by a constant force torsional spring that applies a torque of $145$~mNm to the keyed shaft.

The main component of the clutch unit is the capstan clutch mechanism \cite{In2012}. The mechanism consists of a stainless steel wire wrapped around an aluminum drum, see Fig.~\ref{fig:lock_spring}(b,c). The $2.4$~mm diameter wire is turned $6$ times around the $38$~mm diameter drum of length $20$~mm. The resulting coils are housed in a 3-D printed guide that encases the drum. One end of the wire is terminated in the housing of the clutch unit, and the other end is terminated in the armature of a solenoid (LEDEX 195202-231). Around the armature is a small spring that pre-tensions the stainless steel wire with a small locking force ($F_\text{lock} = 0.65$~N).

\begin{figure*}[ht!]
	\centering
	\includegraphics[width =0.9\textwidth]{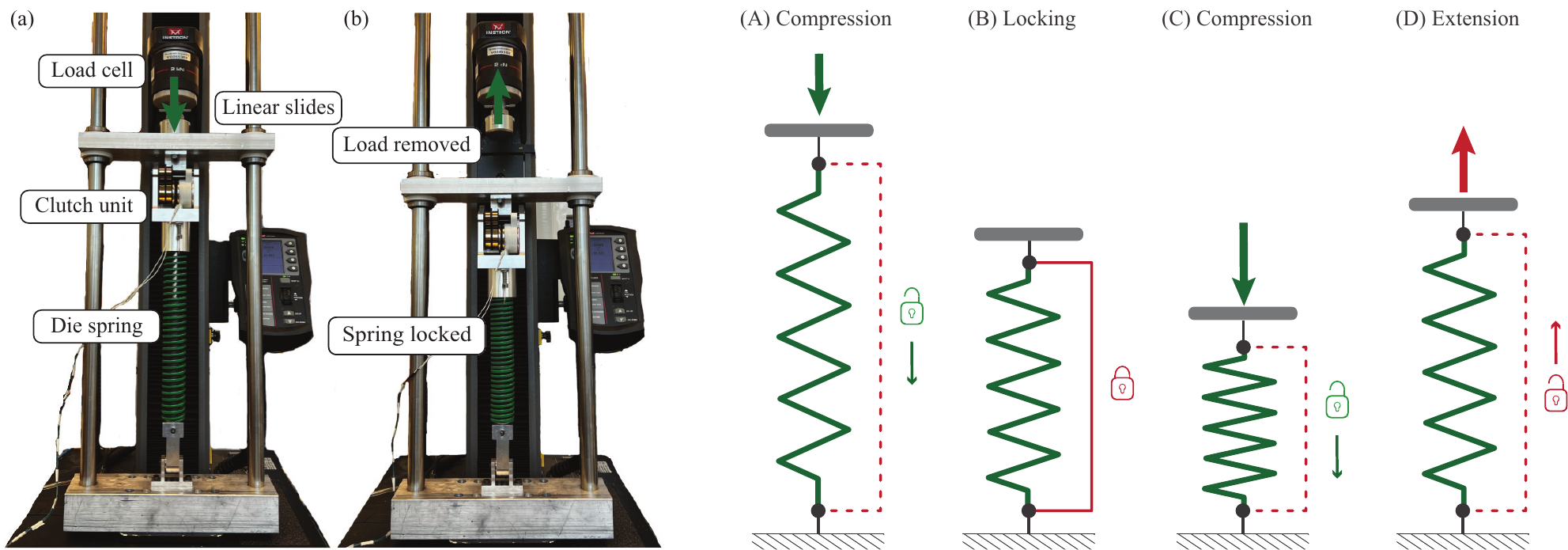}
	\caption{Experimental setup: Instron testing apparatus while (a) the spring being compressed and (b) the spring being locked. Work-loop for mechanical energy accumulation. (A-D) The four sub-figures represent the loading configurations of a lockable spring.}
	\label{fig:exp_setup}
\end{figure*}

\subsection{Locking and unlocking force}
As the spring is compressed, the braided cable that travels through the middle of the spring is wound around a pulley, Fig.~\ref{fig:lock_spring}(c,d). A torsional spring maintains tension in the cable by applying torque about the rotational axis of the pulley, see Fig.~\ref{fig:lock_spring}(d). 
Any attempt to extend the spring and unwind the cable will reverse the rotation of the pulley in the braking direction. When the clutch is braking, the rotation of the pulley is locked. As a result, neither the cable nor the spring can extend, and the energy stored by the spring is retained.

The capstan clutch uses friction to brake in the direction of the wire coil. The braking force is created from the surface contact between the stainless steel wire and the aluminum drum. The friction between the surfaces prevents rotation by tightening the steel wire around the drum when the drum attempts to rotate. Surface contact is maintained by pulling the wire with a small spring placed around the solenoid armature that anchors one end of the wire. The maximal locking force is defined by \cite{Stuart1961}:
\begin{equation}
	\lambda_F = \frac{|F_\text{lock}|}{|F_\text{spring}|} = \frac{r_p}{r_d} e^{-2\pi \mu \frac{l_d}{d_w}}< 0.001,
\end{equation}
where $r_p=12$~mm is the radius of the cable pulley, $r_d=19$~mm is the radius of the aluminum braking drum, $l_d=20$~mm is the width of the braking drum, $d_w=2.4$~mm is the diameter of the stainless steel wire while $\mu=0.4$ is the coefficient of static friction. Retaining elastic potential energy with the capstan does not require control force or external energy, whereas releasing the energy stored by the spring requires the solenoid to compensate the small locking force provided by the tensioning spring, which is at most $F_\text{lock} = 0.65$~N for our device. 

\subsection{Mass-energy density}
The total mass of the lockable spring is $1.94$~kg. The mass includes the die spring, the controllable capstan clutch unit, and the telescoping shafts that prevent buckling. In comparison, the mass of the spring and the support structures, excluding the capstan clutch, is $1.32$~kg. Thus, the ratio between the mass-energy density of the spring with and without the capstan clutch is:
\begin{equation}
	\rho_E =\frac{m_\text{spring}}{m_\text{spring}+m_\text{lock}}= 0.68.
\end{equation}
We find that the mass-energy density of our prototype is comparable to that of the small non-controllable ratchet and pawl mechanism used in \cite{Collins2015} (we calculated $\rho_E=0.63$ by using $0.098$~kg for the spring mass and $0.057$~kg for the mass of the clutch).

We anticipate that, by using weight and material optimization, we could further increase the mass energy density of the prototype presented in this paper. 

\section{Energy Efficiency}\label{sect: exp}
In order to quantify the energy efficiency of the lockable spring, an experiment was conducted. The experimental setup is shown in Fig.~\ref{fig:exp_setup}(a,b). The prototype was configured in a vertically constrained experimental setup, and an Instron 5944 compression testing device (equipped with a $2$~kN load-cell) was used to measure the spring force.

\subsection{Typical work-loop}
In a typical work-loop, the spring is compressed by an external load, and the energy stored by the spring upon compression is retained by the automatic activation of the capstan clutch as the spring starts to extend, see Fig.~\ref{fig:exp_setup}(A,B). The compression and locking may be done repeatedly to accumulate a large amount of energy, see Fig.~\ref{fig:exp_setup}(C). Once the desired amount of energy is accumulated in the locked spring, the energy is released by unlocking the capstan clutch and allowing the spring to extend, see Fig.~\ref{fig:exp_setup}(D).

\subsection{Experimental procedure}
An experiment demonstrating the mechanical energy accumulation capability of the lockable spring was performed. Through this experiment, the functionality of the locking mechanism was tested, and the efficiency of the lockable spring was assessed. The experiment was based on the compression-locking-extension work-loop shown in Fig.~\ref{fig:exp_setup}(A-D). The details of the experiment are introduced below.

During the experiment, the spring was incrementally compressed and locked to five increasing spring deflections, $\Delta l \in\{10,30,50,70,90\}$~mm. To achieve these spring deflections, the following sequential loading procedure was performed. First, the initially un-deflected spring was compressed by $\Delta l =10$~mm, see Fig.~\ref{fig:exp_setup}(A). Second, the Instron testing apparatus was commanded to remove the compressive load by raising until the locked spring lost contact with the load-cell, see Fig.~\ref{fig:exp_setup}(B). Third, the Instron lowered back onto the already compressed spring and compressed it from the previously locked deflection, $\Delta l = 10$~mm, until $\Delta l =30$~mm, see Fig.~\ref{fig:exp_setup}(C). This loading procedure was repeated until the compressed length reached $\Delta l =90$~mm. Subsequently, the capstan clutch was disengaged, and the spring extended to release the energy it stored, see Fig.~\ref{fig:exp_setup}(D). Throughout the entire process, the Instron testing apparatus ensured slow compression and extension of the spring, $0.5$~mm/s in both directions, and the force-deflection data was recorded.

\begin{figure}
	\centering
	\includegraphics[width =\linewidth]{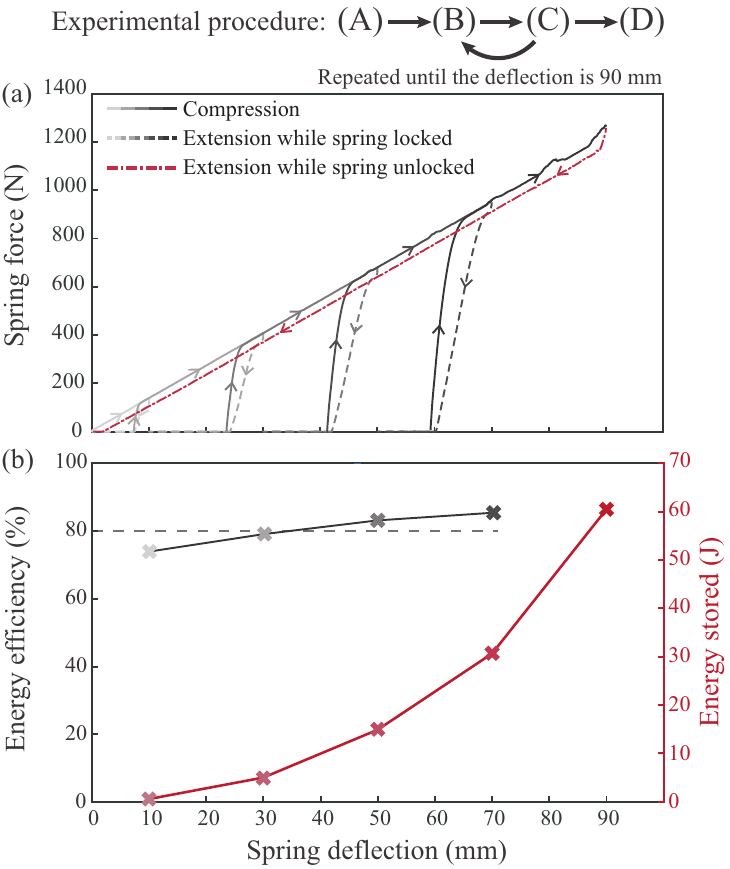}
	\caption{Results of the experiment: (a) Measured force-deflection data. (b) Energy efficiency and the energy stored in the spring.}
	\label{fig:acc_results}
\end{figure}

\subsection{Efficiency of the lockable spring}
Figure~\ref{fig:acc_results}(a) shows the results of the sequential compression test. Figure~\ref{fig:acc_results}(a) (solid line) represents the force of the spring while it is compressed whereas Fig.~\ref{fig:acc_results}(a) (dashed line) represents the force of the spring while it extends. Figure~\ref{fig:acc_results}(b) summarizes the experimental data. Figure~\ref{fig:acc_results}(b) (solid black line) represents the efficiency of the spring after each compression. Figure~\ref{fig:acc_results}(b) (solid red line) represents the total energy stored by the spring.

Based on the experimental data, the capstan clutch was able to lock the spring force up to $1000$~N. Furthermore, the energy stored by the spring was more than $50$~J, while the energy efficiency of the lockable spring was:
	\begin{equation} \label{eq:eta}
		\eta \in [74,84]\%.
	\end{equation}
In order to calculate the energy efficiency, the energy loss is computed by taking the area under the rapidly decreasing portions of the force-deflection curve in Fig.~\ref{fig:acc_results}(a) (dashed lines). The rapidly decreasing portions of the force-deflection curve represent the spring force when the lock is engaged. 
We find that the spring is capable of an average $80\%$ energy storage and return efficiency using the capstan clutch. 

\subsection{Force, time, and energy required to lock the spring}	
Retaining elastic potential energy in a compressed spring did not require control force or external energy. However, releasing the energy stored by the spring required the solenoid to compensate the locking force provided by the small spring located in the capstan lock. The solenoid control was performed off-board with a spike-and-hold method to instantaneously pull in and hold the armature of the solenoid. The control of the capstan lock was done rapidly: it took less than $10$~ms for the brake to fully disengage and release the locked spring providing $1000$~N force. A video demonstrating the fast operation of the capstan clutch is provided in the supplementary material. 

As mentioned previously, the capstan clutch is naturally locked, and, as such, holding over $1000$~N spring force and retaining over $50$~J of energy required no control force or external energy. 
The only energy required to operate the lock was to unlock the capstan clutch. The energy to unlock the clutch was the energy required to move the solenoid against the small locking force $F_\text{lock}=0.65$~N. Consequently, the energy required to unlock the spring was mainly dependent on the locking speed, instead of the large spring force. In order to lock the capstan clutch in $\Delta t=10$~ms, we used $p=15~V \times 0.6~A <10$~W peak power; consequently, the energy required to unlock the clutch was $\Delta E=p\Delta t < 0.1$~J. This amount of energy is less than $0.2\%$ of the energy stored by the spring, see Fig.~\ref{fig:acc_results}(b) (red line).  

An alternative to the capstan clutch is the electro-adhesive clutch \cite{Diller2016} that has lower power consumption ($17$mW), lower weight ($52$~g), but longer response time ($20$~ms) for a comparable $1000$~N holding force. The first two features are desirable, but electro-adhesive clutches also require a high-voltage source to operate, which does not benefit autonomous applications \cite{Diller2018}.

\section{Conclusion}
In this work, we introduced a novel, self-contained design of a compression spring that is continuously lockable and unlockable at high spring forces and exhibits elastic behavior nearly identical to that of a regular compression spring. We showed that the spring is able to efficiently store and hold elastic potential energy at any desired locking position and can return that energy with $20\%$ energy loss. 

The main limitation of the lockable spring is related to the energy loss due to the non-instantaneous locking, the compliance of the cable limiting the deformation of the spring, and the abrasive wear inevitable to capstan mechanisms due to the friction between the components. Designing a capstan clutch with faster self-locking capability and using a cable with high axial stiffness could increase the energy efficiency of lockable springs. Material optimization for the wire rope and the braking drum may mitigate the wear between the components while maintaining the necessary friction coefficient. 

The most comparable existing mechanism to the proposed lockable spring is the clutch-able bi-directional spring recently introduced in \cite{Plooij2016}, because it is also self-contained and can therefore self-lock the spring to hold energy. 
However, the clutch-able bi-directional spring and the prototype lockable spring presented in this paper are developed for different applications. Namely, compared to the lockable spring presented in this paper, the clutch-able bi-directional spring can store less than $2\%$ of the energy (less than $1$~J), has approximately half the mass energy density ($\rho_E<0.32$), and needs at least twice as much time to both lock and unlock the spring ($20$~ms).

Regardless of the design, the ability to retain elastic potential energy in the mechanical domain and release the stored energy at a precisely controlled timing can provide the means for energy efficient spring-based actuation. For example, lockable springs can be applied in pick-and-place robots, where controlling the timing to release the energy stored by the spring may be useful in performing an array of tasks throughout the workspace of the robot, instead of purely cyclic tasks that align with the elastic behavior of a non-lockable spring \cite{Plooij2016}. Additionally, the ability to accumulate energy can provide a means to generate more force for physically demanding tasks.
Lockable springs may be also used in mechanically adaptive, but energetically-conservative, spring mechanisms \cite{Zhang2021,Kim2021}, developed to help robots and humans perform physically demanding tasks using little to no external energy.
\section*{ACKNOWLEDGMENT}
The authors would like to thank Isabel Peppard and John Rector for their assistance during the experimentation.
\bibliographystyle{ieeetr}
\bibliography{bibliography}
\end{document}